%% file: main.tex
\documentclass{article}

\usepackage[nonatbib,preprint]{neurips_2020}

\usepackage[utf8]{inputenc} 
\usepackage[T1]{fontenc}    
\usepackage{hyperref}       
\usepackage{url}            
\usepackage{booktabs}       
\usepackage{amsfonts}       
\usepackage{nicefrac}       
\usepackage{microtype}      

\usepackage{graphicx} 
\usepackage{caption,subcaption}
\usepackage{overpic}
\usepackage{amsmath}
\usepackage{physics}
\usepackage{color}
\usepackage{wrapfig}
\usepackage[numbers,sort&compress]{natbib}
\usepackage{cleveref}
\usepackage[final]{pdfpages}

\usepackage[OT2,T1]{fontenc}
\DeclareSymbolFont{cyrletters}{OT2}{wncyr}{m}{n}
\DeclareMathSymbol{\Sha}{\mathalpha}{cyrletters}{"58}

\DeclareMathOperator*{\argmin}{arg\,min}

\usepackage{xcolor}
\newcommand{\hlw}[1]{{#1}}
\newcommand{\hla}[1]{{#1}}

\title{Implicit Neural Representations with Periodic Activation Functions}

\author{%
	Vincent Sitzmann\thanks{These authors contributed equally to this work.}\\
	\texttt{sitzmann@cs.stanford.edu}
	\And
	Julien N. P. Martel\footnotemark[1]\\
	\texttt{jnmartel@stanford.edu}
	\And
	Alexander W. Bergman\\
	\texttt{awb@stanford.edu}
	\And
	David B. Lindell\\
	\texttt{lindell@stanford.edu}
	\And
	Gordon Wetzstein\\
	\texttt{gordon.wetzstein@stanford.edu}
}

\begin{document}

\maketitle

\vbox{%
	\vskip -0.26in
	\hskip -0.15in
	\hsize\textwidth
	\linewidth\hsize
	\centering
	\normalsize
	Stanford University\\
	\tt\href{https://vsitzmann.github.io/siren/}{vsitzmann.github.io/siren/}
	\vskip 0.3in
}

\input{sections/definitions}

\begin{abstract}
	Implicitly defined, continuous, differentiable signal representations parameterized by neural networks have emerged as a powerful paradigm, offering many possible benefits over conventional representations.
	However, current network architectures for such implicit neural representations are incapable of modeling signals with fine detail, and fail to represent a signal's spatial and temporal derivatives, despite the fact that these are essential to many physical signals defined implicitly as the solution to partial differential equations.
	We propose to leverage periodic activation functions for implicit neural representations and demonstrate that these networks, dubbed sinusoidal representation networks or \sinet{}s, are ideally suited for representing complex natural signals and their derivatives.
	We analyze \sinet{} activation statistics to propose a principled initialization scheme and demonstrate the representation of images, wavefields, video, sound, and their derivatives. 
	Further, we show how \sinet{}s can be leveraged to solve challenging boundary value problems, such as particular Eikonal equations (yielding signed distance functions), the Poisson equation, and the Helmholtz and wave equations.
	Lastly, we combine \sinet{}s with hypernetworks to learn priors over the space of \sinet{} functions.	
	Please see the \href{https://vsitzmann.github.io/siren/}{project website} for a video overview of the proposed method and all applications.
\end{abstract}
	
\section{Introduction}
\label{sec:intro}
\input{sections/introduction}

\section{Related Work}
\label{sec:related_work}
\input{sections/related_work}

\section{Formulation}
\label{sec:representation}
\input{sections/representation}

\begin{figure}[t]
	\includegraphics[width=\textwidth]{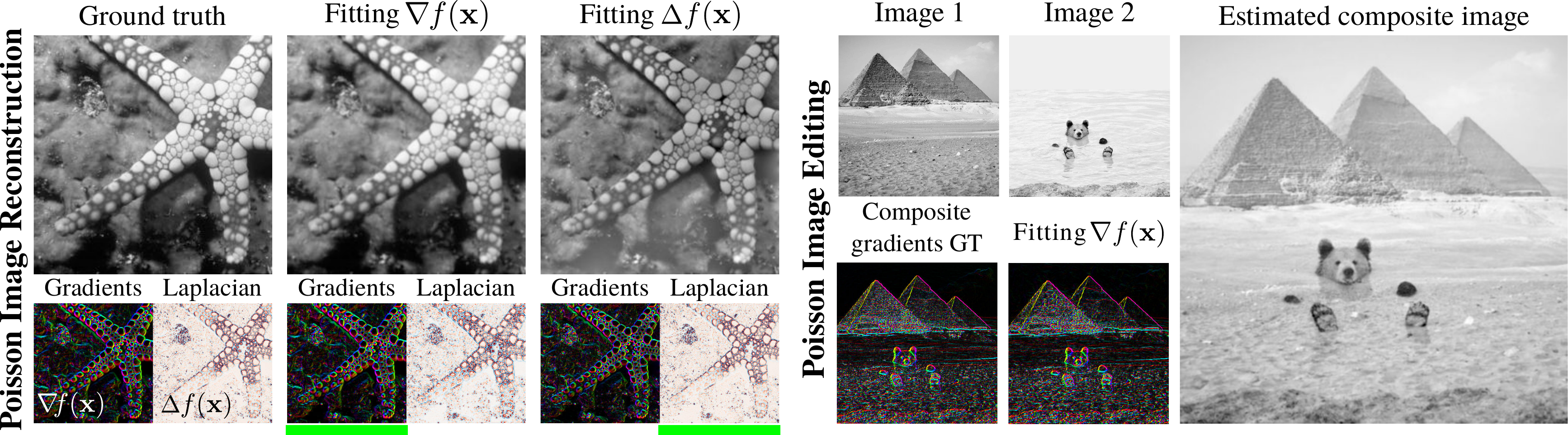}
	\caption{\textbf{Poisson image reconstruction:} An image (left) is reconstructed by fitting a \sinet{}, supervised either by its gradients or Laplacians (underlined in green). The results, shown in the center and right, respectively, match both the image and its derivatives well. \textbf{Poisson image editing:} The gradients of two images (top) are fused (bottom left). \sinet{} allows for the composite (right) to be reconstructed using supervision on the gradients (bottom right).}
	\label{fig:poisson_results}
\end{figure}

\section{Experiments}
In this section, we leverage \sinet{}s to solve challenging boundary value problems using different types of supervision of the derivatives of $\Phi$.
We first solve the Poisson equation via direct supervision of its derivatives.
We then solve a particular form of the Eikonal equation, placing a unit-norm constraint on gradients, parameterizing the class of signed distance functions (SDFs). \sinet{} significantly outperforms ReLU-based SDFs, capturing large scenes at a high level of detail.
We then solve the second-order Helmholtz partial differential equation, and the challenging inverse problem of full-waveform inversion.
Finally, we combine \sinet{}s with hypernetworks, learning a prior over the space of parameterized functions.
All code and data will be made publicly available.

\subsection{Solving the Poisson Equation}
\label{sec:poisson}

\input{sections/poisson}

\subsection{Representing Shapes with Signed Distance Functions}
\label{sec:sdf}
\input{sections/sdf}

\subsection{Solving the Helmholtz and Wave Equations}
\label{sec:applications_bvp}
\input{sections/applications_bvp}

\subsection{Learning a Space of Implicit Functions}
\label{sec:generalization}
\input{sections/generalization}

\section{Discussion and Conclusion}
\label{sec:discussion}
\input{sections/discussion}

\section*{Broader Impact}
\label{sec:bi}
\input{sections/broader_impact}

\begin{ack}
	Vincent Sitzmann, Alexander W. Bergman, and David B. Lindell were supported by a Stanford Graduate Fellowship.
	Julien N. P. Martel was supported by a Swiss National Foundation (SNF) Fellowship (P2EZP2 181817).
	Gordon Wetzstein was supported by an NSF CAREER Award (IIS 1553333), a Sloan Fellowship, and a PECASE from the ARO.
\end{ack}

{\small
	\bibliographystyle{unsrtnat}
	\bibliography{main}
}

\includepdf[pages=-]{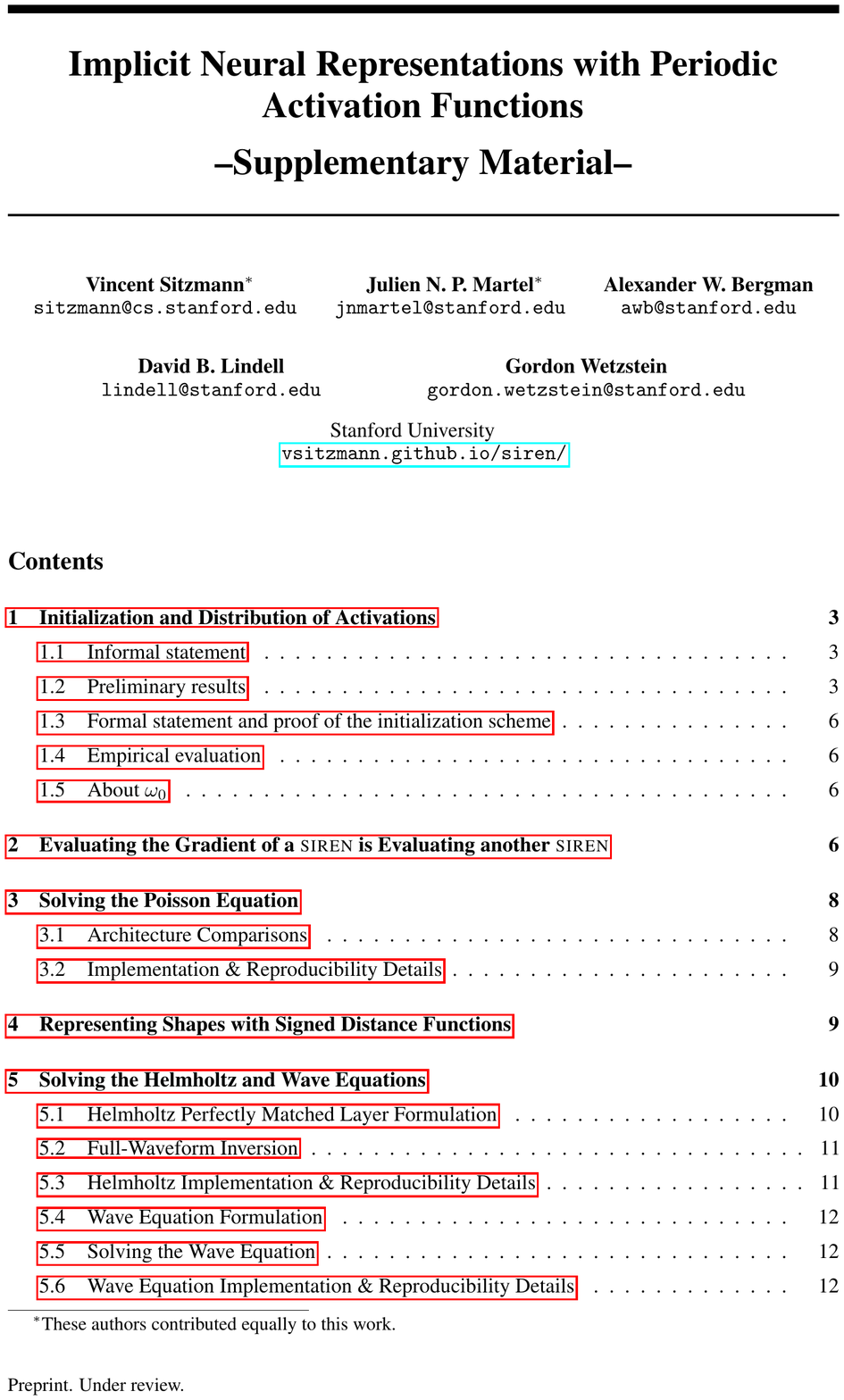}

\end{document}

%% file: sections/definitions.tex
\newcommand{\sinet}{{\sc siren}}

\newcommand{\note}[1]{\textcolor{red}{#1}}

\newcommand{\coords}{\mathbf{x}}
\newcommand{\constants}{\mathbf{a}}
\newcommand{\paramnn}{\theta}

\newcommand{\loss}{\mathcal{L}}

\newcommand{\lapl}{\Delta}
\newcommand{\nablasq}{\grad^2}

\newcommand{\implicitfunction}{F}
\newcommand{\implicitlayer}{\phi}
\newcommand{\implicit}{\Phi}
\newcommand{\implicitparams}{\theta}
\newcommand{\nonlinearlayer}{\sin}

\newcommand{\gt}{f}

\newcommand{\hypernet}{\Psi}
\newcommand{\hypernetparams}{\psi}

\newcommand{\latentcode}{z}

\newcommand{\downsamplingkernel}{h}
\newcommand{\samplingop}{\Sha}
\newcommand{\lowresimg}{b}
\newcommand{\regularizer}{\gamma}

\newcommand{\sdfdomain}{\Omega}
\newcommand{\sdfdomainzero}{\sdfdomain_0}

\newcommand{\context}{O}
\newcommand{\encoder}{C}

%% file: sections/introduction.tex
We are interested in a class of functions $\implicit$ that satisfy equations of the form
\begin{equation}
\implicitfunction \left ( \coords, \implicit, \nabla_\mathbf{x} \implicit, \nabla^2_\mathbf{x} \implicit, \ldots \right) = 0, \quad \Phi: \coords \mapsto \Phi(\coords).
\label{eqn:functional}
\end{equation}
This implicit problem formulation takes as input the spatial or spatio-temporal coordinates $\mathbf{x} \in \mathbb{R}^m$ and, optionally, derivatives of $\implicit$ with respect to these coordinates. 
Our goal is then to learn a neural network that parameterizes $\Phi$ to map $\mathbf{x}$ to some quantity of interest while satisfying the constraint presented in~\Cref{eqn:functional}. Thus, $\Phi$ is implicitly defined by \hla{the relation defined by $F$} and we refer to neural networks that parameterize such implicitly defined functions as \emph{implicit neural representations}.
As we show in this paper, a surprisingly wide variety of problems across scientific fields fall into this form, such as modeling many different types of discrete signals in image, video, and audio processing using a continuous and differentiable representation, learning 3D shape representations via signed distance functions~\cite{park2019deepsdf,mescheder2019occupancy,saito2019pifu,atzmon2019sal}, and, more generally, solving boundary value problems, such as the Poisson, Helmholtz, or wave equations.

\hlw{A continuous parameterization} offers several benefits over alternatives, such as discrete grid-based representations. 
For example, due to the fact that $\Phi$ is defined on the continuous domain of $\coords$, it can be significantly more memory efficient than a discrete representation, allowing it to model fine detail that is not limited by the grid resolution but by the capacity of the underlying network architecture. 
Being differentiable implies that gradients and higher-order derivatives can be computed analytically, for example using automatic differentiation, which again makes these models independent of conventional grid resolutions. 
Finally, with well-behaved derivatives, \hla{implicit neural representations may offer a new toolbox for solving inverse problems, such as differential equations}.

For these reasons, implicit neural representations have seen significant research interest over the last year (Sec.~\ref{sec:related_work}). Most of these recent representations build on ReLU-based multilayer perceptrons (MLPs). While promising, these architectures lack the capacity to represent fine details in the underlying signals, and they typically do not represent the derivatives of a target signal well.
This is partly due to the fact that ReLU networks are piecewise linear, their second derivative is zero everywhere, and they are thus incapable of modeling information contained in higher-order derivatives of natural signals.
While alternative activations, such as tanh or softplus, are capable of representing higher-order derivatives, we demonstrate that their derivatives are often not well behaved and also fail to represent fine details.

To address these limitations, we leverage MLPs with periodic activation functions for implicit neural representations.
We demonstrate that this approach is not only capable of representing details in the signals better than ReLU-MLPs, or positional encoding strategies proposed in concurrent work~\cite{mildenhall2020nerf}, but that these properties also uniquely apply to the derivatives, which is critical for many applications we explore in this paper. 

To summarize, the contributions of our work include:
\begin{itemize}
	\item A continuous implicit neural representation using periodic activation functions that fits \hla{complicated signals}, such as natural images and 3D shapes, and their derivatives robustly.
	\item An initialization scheme for training these representations and validation that distributions of these representations can be learned using hypernetworks.
	\item Demonstration of applications in: image, video, and audio representation; 3D shape reconstruction; solving first-order differential equations that aim at estimating a signal by supervising only with its gradients; and solving second-order differential equations.
\end{itemize}

%% file: sections/related_work.tex
\paragraph{Implicit neural representations.} 
Recent work has demonstrated the potential of fully connected networks as continuous, memory-efficient implicit representations for shape parts \cite{genova2019learning,genova2019deep}, objects~\cite{park2019deepsdf,michalkiewicz2019implicit,atzmon2019sal,gropp2020implicit}, or scenes~\cite{sitzmann2019srns,jiang2020local,peng2020convolutional,chabra2020deep}. 
These representations are typically trained from some form of 3D data as either signed distance functions~\cite{park2019deepsdf,michalkiewicz2019implicit,atzmon2019sal,gropp2020implicit,sitzmann2019srns,jiang2020local,peng2020convolutional} or occupancy networks~\cite{mescheder2019occupancy,chen2019learning}. 
In addition to representing shape, some of these models have been extended to also encode object appearance~\cite{saito2019pifu,sitzmann2019srns,Oechsle2019ICCV,Niemeyer2020CVPR,mildenhall2020nerf}, which can be trained using (multiview) 2D image data using neural rendering~\cite{tewari2020state}.
Temporally aware extensions ~\cite{Niemeyer2019ICCV} and variants that add part-level semantic segmentation~\cite{kohli2020inferring} have also been proposed. 
%

\paragraph{Periodic nonlinearities.} 
Periodic nonlinearities have been investigated repeatedly over the past decades, but have so far failed to robustly outperform alternative activation functions.
Early work includes Fourier neural networks, engineered to mimic the Fourier transform via single-hidden-layer networks~\cite{Gallant:1988,zhumekenov2019fourier}.
Other work explores neural networks with periodic activations for simple classification tasks~\cite{sopena1999neural,wong2002handwritten,parascandolo2016taming} and recurrent neural networks~\cite{liu2015multistability,koplon1997using,choueiki1997implementing,alquezar1997symbolic,sopena1994improvement}.
It has been shown that such models have universal function approximation properties~\cite{candes1999harmonic,lin2013approximation,sonoda2017neural}.
Compositional pattern producing networks~\cite{stanley2007compositional,mordvintsev2018differentiable} also leverage periodic nonlinearities, but rely on a combination of different nonlinearities via evolution in a genetic algorithm framework.
Motivated by the discrete cosine transform, \citet{klocek2019hypernetwork} leverage cosine activation functions for image representation but they do not study the derivatives of these representations or other applications explored in our work.
Inspired by these and other seminal works, we explore MLPs with periodic activation functions for applications involving implicit neural representations and their derivatives, and we propose principled initialization and generalization schemes.

\paragraph{Neural DE Solvers.} 
Neural networks have long been investigated in the context of solving differential equations (DEs)~\cite{lee1990neural}, and have previously been introduced as implicit representations for this task~\cite{lagaris1998artificial}.
Early work on this topic involved simple neural network models, consisting of MLPs or radial basis function networks with few hidden layers and hyperbolic tangent or sigmoid nonlinearities~\cite{lagaris1998artificial,he2000multilayer,mai2003approximation}.
The limited capacity of these shallow networks typically constrained results to 1D solutions or simple 2D surfaces.
Modern approaches to these techniques leverage recent optimization frameworks and auto-differentiation, but use similar architectures based on MLPs.
Still, solving more sophisticated equations with higher dimensionality, more constraints, or more complex geometries is feasible~\cite{sirignano2018dgm,raissi2019physics,berg2018unified}.
However, we show that the commonly used MLPs with smooth, non-periodic activation functions fail to accurately model high-frequency information and higher-order derivatives even with dense supervision.

Neural ODEs~\cite{chen2018neural} are related to this topic, but are very different in nature.
Whereas implicit neural representations can be used to directly solve ODEs or PDEs from supervision on the system dynamics, neural ODEs allow for continuous function modeling by pairing a conventional ODE solver (e.g., implicit Adams or Runge-Kutta) with a network that parameterizes the dynamics of a function. The proposed architecture may be complementary to this line of work.

%% file: sections/representation.tex
\begin{figure}[t!]
\centering
\includegraphics[width=0.95\linewidth]{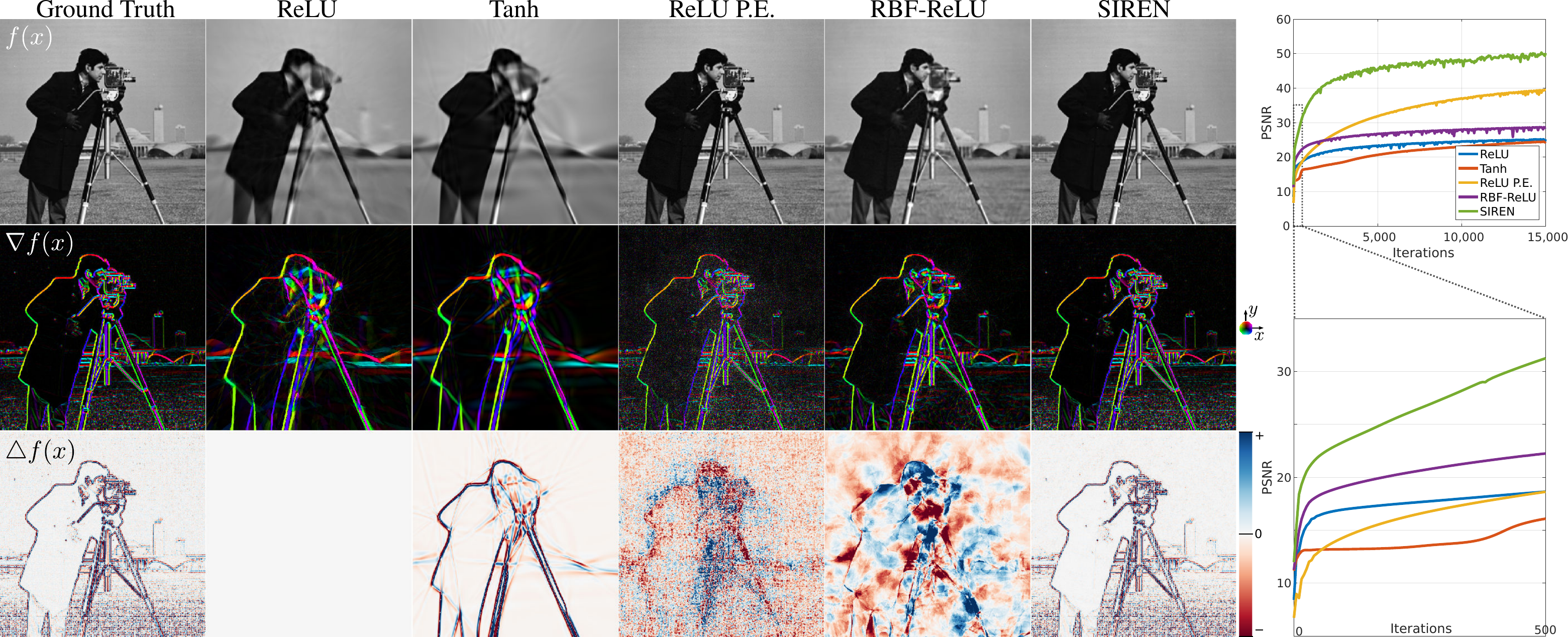}
\caption{Comparison of different implicit network architectures fitting a ground truth image (top left). The representation is only supervised on the target image but we also show first- and second-order derivatives of the function fit in rows~2 and~3, respectively.}
\label{fig:imagefit}
\vspace{-0.22cm}
\end{figure}

Our goal is to solve problems of the form presented in ~\Cref{eqn:functional}.
We cast this as a feasibility problem, where a function $\Phi$ is sought that satisfies a set of $M$ constraints $\{\mathcal{C}_m(\constants(\coords), \Phi(\coords), \nabla\Phi(\coords), ...)\}_{m=1}^M$, each of which relate the function $\Phi$ and/or its derivatives to quantities $\constants(\mathbf{x})$:
\begin{equation}
\text{find} \: \Phi(\coords) \quad{}
\text{subject to} \:\, \mathcal{C}_m\big(\constants(\coords), \Phi(\coords), \nabla\Phi(\coords), ...\big) = 0, \: \forall \coords\in\Omega_m, \,  m = 1, \ldots, M
\end{equation}
This problem can be cast in a loss function that penalizes deviations from each of the constraints on their domain $\Omega_m$:
\begin{equation}
\mathcal{L} = \int_\Omega \sum_{m=1}^M \mathbf{1}_{\Omega_m}(\coords)\, \| \mathcal{C}_m(\mathbf{a}(\coords), \Phi(\coords), \nabla\Phi(\coords), ...)\| d\coords,
\label{eq:general_cont_loss_function}
\end{equation}
with the indicator function $\mathbf{1}_{\Omega_m}(\coords)=1$ when $\coords\in\Omega_m$ and $0$ when $\coords \not\in \Omega_m$. In practice, the loss function is enforced by sampling $\Omega$. A dataset $\mathcal{D} = \{ (\coords_i, \mathbf{a}_i(\coords)) \}_i$ is a set of tuples of coordinates $\coords_i\in\Omega$ along with samples from the quantities $\mathbf{a}(\coords_i)$ that appear in the constraints. 
Thus, the loss in Equation~\eqref{eq:general_cont_loss_function} is enforced on coordinates $\coords_i$ sampled from the dataset, yielding the loss
$\tilde{\mathcal{L}}=\sum_{i\in\mathcal{D}} \sum_{m=1}^M \| \mathcal{C}_m(a(\coords_i), \Phi(\coords_i), \nabla\Phi(\coords_i), ...)\|$. In practice, the dataset $\mathcal{D}$ is sampled dynamically at training time, approximating $\mathcal{L}$ better as the number of samples grows, as in Monte Carlo integration.

We parameterize functions $\Phi_{\paramnn}$ as fully connected neural networks with parameters $\paramnn$, and solve the resulting optimization problem using gradient descent.

\subsection{Periodic Activations for Implicit Neural Representations}
We propose \sinet{}, a simple neural network architecture for implicit neural representations that uses the sine as a periodic activation function:
\begin{equation}
\implicit \left( \mathbf{x} \right) = \mathbf{W}_n \left( \implicitlayer_{n-1} \circ \implicitlayer_{n-2} \circ \ldots \circ \implicitlayer_0 \right) \left( \mathbf{x} \right) + \mathbf{b}_n, \quad \mathbf{x}_i \mapsto \implicitlayer_i \left( \mathbf{x}_i \right) = \nonlinearlayer \left( \mathbf{W}_i \mathbf{x}_i + \mathbf{b}_i \right).
\end{equation}
Here, $\implicitlayer_i: \mathbb{R}^{M_i} \mapsto \mathbb{R}^{N_i}$ is the $i^{th}$ layer of the network. It consists of the affine transform defined by the weight matrix $\mathbf{W}_i \in \mathbb{R}^{N_i \times M_i}$ and the biases $\mathbf{b}_i\in  \mathbb{R}^{N_i}$ applied on the input $\coords_i\in\mathbb{R}^{M_i}$, followed by the sine nonlinearity applied to each component of the resulting vector.

Interestingly, any derivative of a \sinet{} \emph{is itself a \sinet{}}, as the derivative of the sine is a cosine, i.e., a phase-shifted sine (see supplemental). 
Therefore, the derivatives of a \sinet{} inherit the properties of \sinet{}s, enabling us to supervise any derivative of \sinet{} with  ``complicated'' signals.
In our experiments, we demonstrate that when a \sinet{} is supervised using a constraint $\mathcal{C}_m$ involving the derivatives of $\phi$, the function $\phi$ remains well behaved, which is crucial in solving many problems, including boundary value problems (BVPs).

We will show that \sinet{}s can be initialized with some control over the distribution of activations, allowing us to create deep architectures. Furthermore, \sinet{}s converge significantly faster than baseline architectures, fitting, for instance, a single image in a few hundred iterations, taking a few seconds on a modern GPU, while featuring higher image fidelity (Fig.~\ref{fig:imagefit}).

\begin{wrapfigure}[16]{r}{0.6\textwidth}
	\vspace{-0.4cm}
	\includegraphics[width=0.58\textwidth]{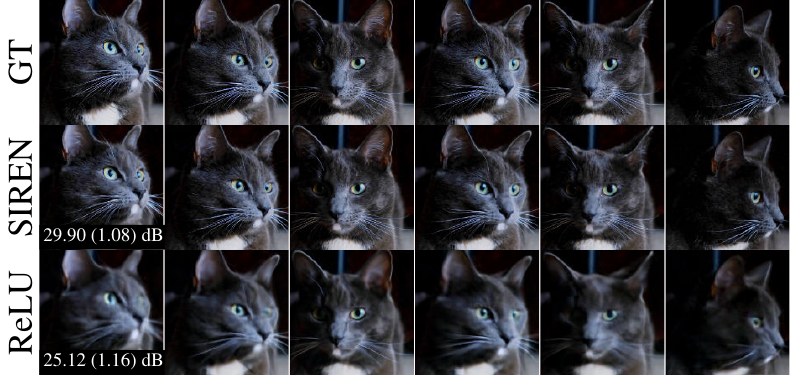}
	\caption{Example frames from fitting a video with \sinet{} and ReLU-MLPs. Our approach faithfully reconstructs fine details like the whiskers. Mean (and standard deviation) of the PSNR over all frames is reported.}
	\label{fig:cat}
\end{wrapfigure}

\paragraph{A simple example: fitting an image.} 
Consider the case of finding the function $\implicit:\mathbb{R}^2 \mapsto \mathbb{R}^3, \coords \to \implicit(\coords)$ that parameterizes a given discrete image $f$ in a continuous fashion. The image defines a dataset $\mathcal{D}=\{(\coords_{i}, f(\coords_i))\}_i$ of pixel coordinates $\coords_i=(x_i,y_i)$ associated with their RGB colors $f(\coords_i)$. The only constraint $\mathcal{C}$ enforces is that $\implicit$ shall output image colors at pixel coordinates, solely depending on $\implicit$ (none of its derivatives) and $f(\coords_i)$, with the form $\mathcal{C}(f(\coords_i),\implicit(\coords))=\implicit(\coords_i) - f(\coords_i)$ which can be translated into the loss $\tilde{\mathcal{L}} = \sum_{i} \| \implicit(\coords_i) - f(\coords_i)\|^2$.
In Fig.~\ref{fig:imagefit}, we fit $\implicit_\theta$ using comparable network architectures with different activation functions to a natural image.
We supervise this experiment only on the image values, but also visualize the gradients $\grad\gt$ and Laplacians $\Delta\gt$.
While only two approaches, a ReLU network with positional encoding (P.E.)~\cite{mildenhall2020nerf} and our \sinet{}, accurately represent the ground truth image $\gt \left( \mathbf{x} \right)$, \sinet{} is the only network capable of also representing the derivatives of the signal.
Additionally, we run a simple experiment where we fit a short video with 300 frames and with a resolution of 512$\times$512 pixels using both ReLU and \sinet{} MLPs. As seen in Figure~\ref{fig:cat}, our approach is successful in representing this video with an average peak signal-to-noise ratio close to 30~dB, outperforming the ReLU baseline by about 5~dB.
We also show the flexibility of \sinet{}s by representing audio signals in the supplement.

\subsection{Distribution of activations, frequencies, and a principled initialization scheme}
We present a principled initialization scheme necessary for the effective training of \sinet{}s. 
While presented informally here, we discuss further details, proofs and empirical validation in the supplemental material. 
The key idea in our initialization scheme is to preserve the distribution of activations through the network so that the final output at initialization does not depend on the number of layers. Note that building \sinet{}s with not carefully chosen uniformly distributed weights yielded poor performance both in accuracy and in convergence speed. 

To this end, let us first consider the output distribution of a single sine neuron with the uniformly distributed input $x \sim \mathcal{U}(-1, 1)$. The neuron's output is $y = \sin(ax + b)$ with $a,b \in\mathbb{R}$.
It can be shown that for any $a>\frac{\pi}{2}$, i.e. spanning at least half a period, the output of the sine is $y\sim\text{arcsine}(-1,1)$, a special case of a U-shaped Beta distribution and independent of the choice of $b$. 
We can now reason about the output distribution of a neuron. Taking the linear combination of $n$ inputs $\mathbf{x}\in\mathbb{R}^n$ weighted by $\mathbf{w}\in\mathbb{R}^n$, its output is $y=\sin(\mathbf{w}^T\mathbf{x} + b)$.
Assuming this neuron is in the second layer, each of its inputs is arcsine distributed. When each component of $\mathbf{w}$ is uniformly distributed such as $w_i \sim \mathcal{U}(-c/{\sqrt{n}}, c/{\sqrt{n}}),c\in\mathbb{R}$, we show (see supplemental) that the dot product converges to the normal distribution $\mathbf{w}^T\mathbf{x} \sim \mathcal{N}(0, c^2/6)$ as $n$ grows.
Finally, feeding this normally distributed dot product through another sine is also arcsine distributed for any $c>\sqrt{6}$. 
Note that the weights of a \sinet{} can be interpreted as angular frequencies while the biases are phase offsets. Thus, larger frequencies appear in the networks for weights with larger magnitudes.
For $|\mathbf{w}^T\mathbf{x}|<{\pi}/{4}$, the sine layer will leave the frequencies unchanged, as the sine is approximately linear. 
In fact, we empirically find that a sine layer keeps spatial frequencies approximately constant for amplitudes such as $|\mathbf{w}^T\mathbf{x}|<\pi$, and increases spatial frequencies for amplitudes above this value\footnote{Formalizing the distribution of output frequencies throughout \sinet{}s proves to be a hard task and is out of the scope of this work.}.

Hence, we propose to draw weights with $c=6$ so that $w_i \sim \mathcal{U}(-\sqrt{6/n}, \sqrt{6/n})$. 
This ensures that the input to each sine activation is normal distributed with a standard deviation of $1$.
Since only a few weights have a magnitude larger than $\pi$, the frequency throughout the sine network grows only slowly. 
Finally, we propose to initialize the first layer of the sine network with weights so that the sine function $\sin(\omega_0\cdot\mathbf{W}\coords + \mathbf{b})$ spans multiple periods over $[-1,1]$. We found $\omega_0=30$ to work well for all the applications in this work.
The proposed initialization scheme yielded fast and robust convergence using the ADAM optimizer for all experiments in this work.

%% file: sections/poisson.tex
We demonstrate that the proposed representation is not only able to accurately
represent a function and its derivatives, but that it can also be supervised solely
by its derivatives, i.e., the model is never presented with the actual
function values, but only values of its first or higher-order derivatives. 

An intuitive example representing this class of problems is the Poisson equation. The Poisson equation is perhaps the simplest elliptic partial differential equation (PDE) which is crucial in physics and engineering, for example to model potentials arising from distributions of charges or masses. In this problem, an unknown ground truth signal $\gt$ is estimated from discrete samples of either its gradients $\grad \gt$ or Laplacian $\lapl \gt = \grad \cdot \grad \gt$ as 
\begin{equation}
	\loss_{\mathrm{grad.}} = \int_\sdfdomain
	\lVert \grad_{\mathbf{x}}\implicit (\mathbf{x}) - \grad_{\mathbf{x}} \gt (\mathbf{x}) \rVert \,\, d \mathbf{x}, \quad \text{or} \quad 	
	\loss_{\mathrm{lapl.}} = \int_\sdfdomain
	\lVert \lapl \implicit (\mathbf{x}) - \lapl \gt (\mathbf{x}) \rVert \,\, d \mathbf{x}.
	\label{eq:functional_gradient}
\end{equation}

\paragraph{Poisson image reconstruction.}
Solving the Poisson equation enables the reconstruction of images from their derivatives. We show results of this approach using \sinet{} in Fig.~\ref{fig:poisson_results}.
Supervising the implicit representation with either ground truth gradients via $\loss_{\mathrm{grad.}}$ or Laplacians via $\loss_{\mathrm{lapl.}}$ successfully reconstructs the image. Remaining intensity variations are due to the ill-posedness of the problem.

\paragraph{Poisson image editing.}
Images can be seamlessly fused in the gradient domain~\cite{perez:2003}. For this purpose, $\implicit$ is supervised using $\loss_{\mathrm{grad.}}$ of Eq.~\eqref{eq:functional_gradient}, where $\grad_{\mathbf{x}} \gt (\mathbf{x})$ is a composite function of the gradients of two images $\gt_{1,2}$: $\grad_{\mathbf{x}} \gt (\mathbf{x})=\alpha\cdot\grad{\gt_1}(x) + (1-\alpha)\cdot\grad{\gt_2}(x),\:\alpha\in[0,1]$. Fig.~\ref{fig:poisson_results} shows two images seamlessly fused with this approach.

%% file: sections/sdf.tex
\begin{figure}[t!]
	\includegraphics[width=\textwidth]{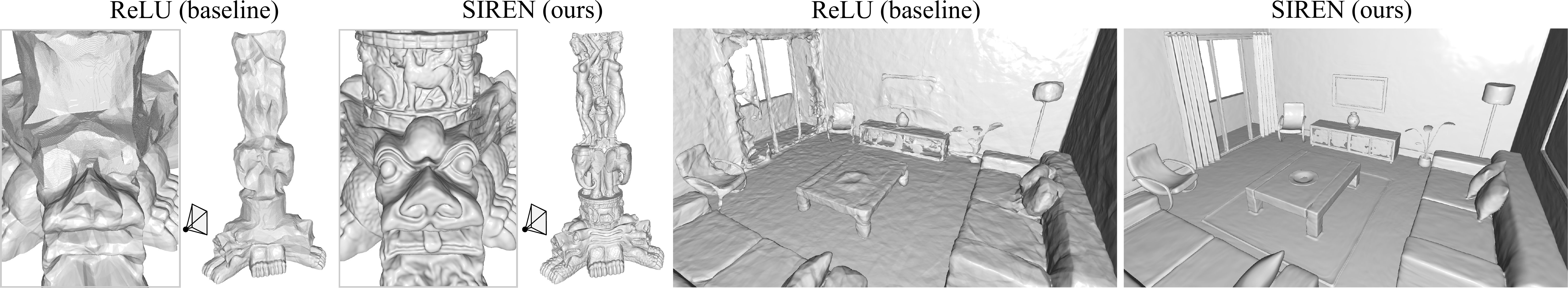}
	\caption{Shape representation. We fit signed distance functions parameterized by implicit neural representations directly on point clouds. Compared to ReLU implicit representations, our periodic activations significantly improve detail of objects (left) and complexity of entire scenes (right).}
	\label{fig:sdf_results}
\end{figure}
Inspired by recent work on shape representation with differentiable signed distance functions (SDFs)~\cite{park2019deepsdf,atzmon2019sal,gropp2020implicit}, we fit SDFs directly on oriented point clouds using both ReLU-based implicit neural representations and \sinet{}s.
This amounts to solving a particular Eikonal boundary value problem that constrains the norm of spatial gradients $
|\nabla_\coords \Phi|$ to be $1$ almost everywhere.
Note that ReLU networks are seemingly ideal for representing SDFs, as their gradients are locally constant and their second derivatives are 0. Adequate training procedures for working directly with point clouds were described in prior work~\cite{atzmon2019sal,gropp2020implicit}.
We fit a \sinet{} to an oriented point cloud using a loss of the form
\begin{equation}
	\mathcal{L}_{\mathrm{sdf}} \! = \! \int_{\sdfdomain}
	\big\| \left|\grad_\mathbf{x} \implicit(\mathbf{x}) \right| - 1 \big\|
	d \mathbf{x}
	+
	\int_{\sdfdomainzero}
	\!\!
		  \left\| \implicit ( \mathbf{x} ) \right\|
	 	+ \big(1 - \langle \grad_\mathbf{x} \implicit(\mathbf{x}), \mathbf{n}(\mathbf{x}) \rangle  \big)
	d \mathbf{x}
	+
	\int_{\sdfdomain\setminus\sdfdomainzero} \!\!\!\!
	\psi \big(\implicit(\mathbf{x})\big)
	d \mathbf{x},
\end{equation}
Here, $\psi(\mathbf{x})=\exp(-\alpha\cdot|\implicit(\mathbf{x})|),\alpha\gg 1$ penalizes off-surface points for creating SDF values close to 0. $\sdfdomain$ is the whole domain and we denote the zero-level set of the SDF as $\sdfdomainzero$. The model $\implicit(x)$ is supervised using oriented points sampled on a mesh, where we require the \sinet{} to respect $\implicit(\mathbf{x})=0$ and its normals $\mathbf{n}(\mathbf{x})=\grad \gt(\mathbf{x})$. During training, each minibatch contains an equal number of points on and off the mesh, each one randomly sampled over $\sdfdomain$.
As seen in Fig.~\ref{fig:sdf_results}, the proposed periodic activations significantly increase the details of objects and the complexity of scenes that can be represented by these neural SDFs, parameterizing a full room with only a single five-layer fully connected neural network.
This is in contrast to concurrent work that addresses the same failure of conventional MLP architectures to represent complex or large scenes by locally decoding a discrete representation, such as a voxelgrid, into an implicit neural representation of geometry~\cite{chabra2020deep,peng2020convolutional,jiang2020local}.

%% file: sections/applications_bvp.tex
\begin{figure}[t]
    \centering
        \hspace{0.12in}
        \includegraphics[]{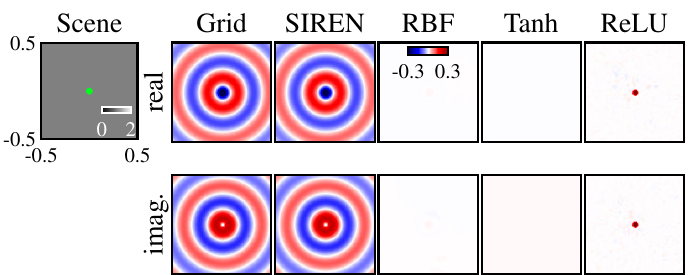}
        \hspace{0.12in}
        \includegraphics[]{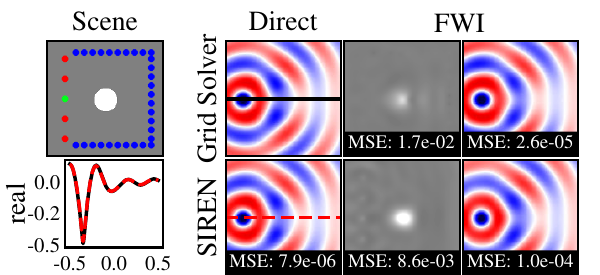}
        \put(-396,8){\rotatebox{90}{\bf{\fontsize{8}{10}\selectfont Direct Inversion}}}
        \put(-182,13){\rotatebox{90}{\bf{\fontsize{8}{10}\selectfont Neural FWI}}}
        \caption{\textbf{Direct Inversion:} We solve the Helmholtz equation for a single point source placed at the center of a medium (green dot) with uniform wave propagation velocity (top left). The \sinet{} solution closely matches a principled grid solver~\cite{chen2013optimal} while other network architectures fail to find the correct solution.
        \textbf{Neural Full-Waveform Inversion (FWI):} A scene contains a source (green) and a circular wave velocity perturbation centered at the origin (top left). With the scene velocity known \textit{a priori}, \sinet{} directly reconstructs a wavefield that closely matches a principled grid solver~\cite{chen2013optimal} (bottom left, middle left). For FWI, the velocity and wavefields are reconstructed with receiver measurements (blue dots) from sources triggered in sequence (green, red dots). The \sinet{} velocity model outperforms a principled FWI solver~\cite{aghamiry2019improving}, accurately predicting wavefields. FWI MSE values are calculated across all wavefields and the visualized real wavefield corresponds to the green source.}
\label{fig:helmholtz_comparison}
\end{figure}

The Helmholtz and wave equations are second-order partial differential equations related to the physical modeling of diffusion and waves.  
They are closely related through a Fourier-transform relationship, with the Helmholtz equation given as  
\begin{equation}
     H(m)\,\implicit(\mathbf{x}) = -\gt(\mathbf{x}), \, 
     \text{with}\: H(m) = \big(\lapl + m(\mathbf{x})\,w^2\big).
    \label{eqn:helmholtz}
\end{equation}
Here, $\gt(\mathbf{x})$ represents a known source function, $\implicit(\mathbf{x})$ is the unknown
wavefield, and the squared slowness $m(\mathbf{x}) = 1/c(\mathbf{x})^2$ is a function of the wave
velocity $c(\mathbf{x})$.
In general, the solutions to the Helmholtz equation are complex-valued and require numerical solvers to compute. As the Helmholtz and wave equations follow a similar form, we discuss the Helmholtz equation here, with additional results and discussion for the wave equation in the supplement.  

\paragraph{Solving for the wavefield.}
We solve for the wavefield by parameterizing $\implicit(\mathbf{x})$ with a \sinet{}.
To accommodate a complex-valued solution, we configure the network to output two values, interpreted as the real and imaginary parts.
Training is performed on randomly sampled points $\mathbf{x}$ within the domain $\Omega = \{\mathbf{x} \in \mathbb{R}^2 \, | \, \lVert \mathbf{x} \rVert_\infty < 1\}$.
The network is supervised using a loss function based on the Helmholtz equation,
$
\mathcal{L}_{\text{Helmholtz}} = \int_\Omega \lambda(\mathbf{x})\,\lVert H(m)\implicit(\mathbf{x}) + \gt(\mathbf{x}) \rVert_1 \,d\mathbf{x},
$
with $\lambda(\mathbf{x})=k$, a hyperparameter, when $f(\mathbf{x})\neq 0$ (corresponding to the inhomogeneous contribution to the Helmholtz equation) and $\lambda(\mathbf{x})=1$ otherwise (for the homogenous part).
Each minibatch contains samples from both contributions and $k$ is set so the losses are approximately equal at the beginning of training. In practice, we use a slightly modified form of Equation (\ref{eqn:helmholtz}) to include the perfectly matched boundary conditions that are necessary to ensure a unique solution~\cite{chen2013optimal} (see supplement for details).

Results are shown in Fig.~\ref{fig:helmholtz_comparison} for solving the Helmholtz equation in two dimensions with spatially uniform wave velocity and a single point source (modeled as a Gaussian with $\sigma^2 = 10^{-4}$).
The \sinet{} solution is compared with a principled solver~\cite{chen2013optimal} as well as other neural network solvers.
All evaluated network architectures use the same number of hidden layers as \sinet{} but with different activation functions.
In the case of the RBF network, we prepend an RBF layer with 1024 hidden units and use a tanh activation.
\sinet{} is the only representation capable of producing a high-fidelity reconstruction of the wavefield.
We also note that the tanh network has a similar architecture to recent work on neural PDE solvers~\cite{raissi2019physics}, except we increase the network size to match \sinet{}.

\paragraph{Neural full-waveform inversion (FWI).}

In many wave-based sensing modalities (radar, sonar, seismic imaging, etc.), one attempts to probe and sense across an entire domain using sparsely placed sources (i.e., transmitters) and receivers.
FWI uses the known locations of sources and receivers to jointly recover the entire wavefield and other physical properties, such as permittivity, density, or wave velocity.
Specifically, the FWI problem can be described as~\cite{van2013mitigating}
\begin{equation}
    \argmin_{m, \Phi} \sum\limits_{i=1}^{N} \int_\Omega \lvert \Sha_r (\implicit_i(\mathbf{x}) - r_i(\mathbf{x})) \rvert^2\, d\mathbf{x} \; \text{s.t.}\;  H(m)\,\implicit_i(\mathbf{x}) = -\gt_i(x), \: 1\le i \le N, \,\forall \mathbf{x} \in \Omega,
    \label{eqn:fwi}
\end{equation}
where there are $N$ sources, $\Sha_r$ samples the wavefield at the receiver locations, and $r_i(x)$ models receiver data for the $i$th source.

We first use a \sinet{} to directly solve Eq.~\ref{eqn:helmholtz} for a known wave velocity perturbation, obtaining an accurate wavefield that closely matches that of a principled solver~\cite{chen2013optimal} (see Fig.~\ref{fig:helmholtz_comparison}, right).
Without \textit{a priori} knowledge of the velocity field, FWI is used to jointly recover the wavefields and velocity. Here, we use 5 sources and place 30 receivers around the domain, as shown in Fig.~\ref{fig:helmholtz_comparison}.
Using the principled solver, we simulate the receiver measurements for the 5 wavefields (one for each source) at a single frequency of 3.2 Hz, which is chosen to be relatively low for improved convergence.
We pre-train \sinet{} to output 5 complex wavefields and a squared slowness value for a uniform velocity.
Then, we optimize for the wavefields and squared slowness using a penalty method variation~\cite{van2013mitigating} of Eq.~\ref{eqn:fwi} (see the supplement for additional details).
In Fig.~\ref{fig:helmholtz_comparison}, we compare to an FWI solver based on the alternating direction method of multipliers~\cite{boyd2011distributed,aghamiry2019improving}.
With only a single frequency for the inversion, the principled solver is prone to converge to a poor solution for the velocity.
As shown in Fig.~\ref{fig:helmholtz_comparison}, \sinet{} converges to a better velocity solution and accurate solutions for the wavefields.
All reconstructions are performed or shown at $256\times 256$ resolution to avoid noticeable stair-stepping artifacts in the circular velocity perturbation.

%% file: sections/generalization.tex
\begin{figure}
	\includegraphics[width=\textwidth]{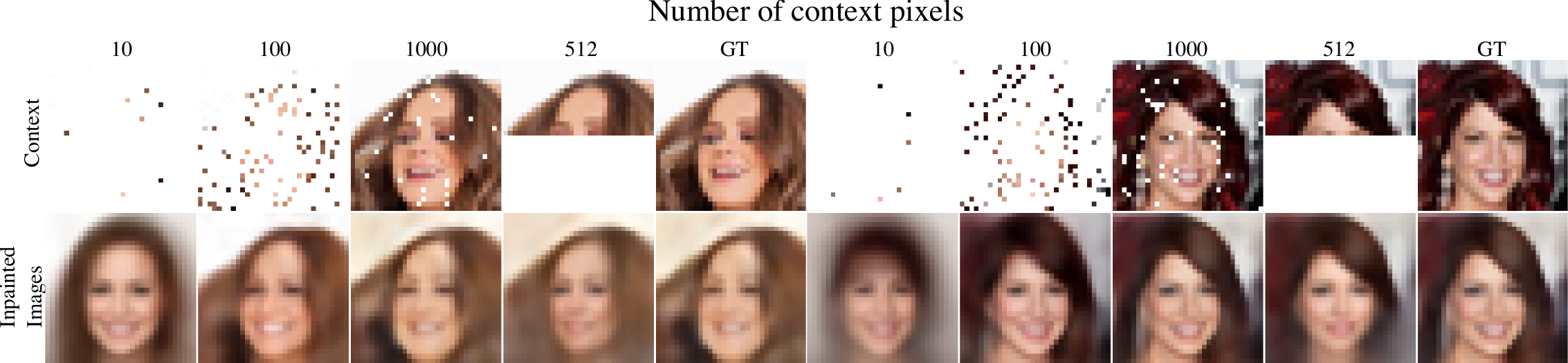} 
	\caption{Generalizing across implicit functions parameterized by \sinet{}s on the CelebA dataset~\cite{liu2015faceattributes}.  
	Image inpainting results are shown for various numbers of context pixels in $\context_j$.}
	\label{fig:generalization}
\end{figure}

A powerful concept that has emerged for implicit representations is to learn priors over the space of functions that define them~\cite{park2019deepsdf,mescheder2019occupancy,sitzmann2019srns}.
Here we demonstrate that the function space parameterized by \sinet{}s also admits the learning of powerful priors. 
Each of these \sinet{}s $\implicit_j$ are fully defined by their parameters $\boldsymbol{\implicitparams}_j \in \mathbb{R}^l$. Assuming that all parameters $\boldsymbol{\implicitparams}_j$ of a class exist in a $k$-dimensional subspace of $\mathbb{R}^l$, $k<l$, then these parameters can be well modeled by latent code vectors in $\mathbf{\latentcode} \in \mathbb{R}^k$. Like in neural processes~\cite{garnelo2018conditional, eslami2018neural, kim2019attentive}, we condition these latent code vectors on partial observations of the signal $\context\in\mathbb{R}^m$ through an encoder
\begin{equation}
\encoder : \mathbb{R}^m \to \mathbb{R}^k, \quad \context_j \mapsto \encoder(\context_j) = \mathbf{\latentcode}_j,
\end{equation}
and use a ReLU hypernetwork~\cite{ha2016hypernetworks}, to map the latent code to the weights of a \sinet{}, as in~\cite{sitzmann2019srns}: 
\begin{equation}
\hypernet : \mathbb{R}^k \to \mathbb{R}^l, \quad \mathbf{\latentcode}_j \mapsto \hypernet(\mathbf{\latentcode}_j) = \boldsymbol{\implicitparams_j}.
\end{equation}

We replicated the experiment from ~\cite{garnelo2018conditional} on the CelebA dataset~\cite{liu2015faceattributes} using a set encoder. Additionally, we show results using a convolutional neural network encoder which operates on sparse images. Interestingly, this improves the quantitative and qualitative performance on the inpainting task.
\begin{wraptable}[10]{r}{0.55\textwidth}
	\caption{Quantitative comparison to Conditional Neural Processes~\cite{garnelo2018conditional} (CNPs) on the $32\times32$ CelebA test set. Metrics are reported in pixel-wise mean squared error.}
	\label{tab:generalization}
	\centering
	\begin{tabular}{lccc}
		\toprule
		Number of Context Pixels & 10 & 100 & 1000 \\
		\midrule
		CNP~\cite{garnelo2018conditional} & 0.039 & 0.016 & 0.009 \\
		Set Encoder + Hypernet. & 0.035 & 0.013 & 0.009 \\
		CNN Encoder + Hypernet. & \textbf{0.033} & \textbf{0.009} & \textbf{0.008} \\
		\bottomrule
	\end{tabular}
\end{wraptable}

At test time, this enables reconstruction from sparse pixel observations, and, thereby, inpainting.
Fig.~\ref{fig:generalization} shows test-time reconstructions from a varying number of pixel observations. Note that these inpainting results were all generated using the same model, with the same parameter values. 
Tab.~\ref{tab:generalization} reports a quantitative comparison to ~\cite{garnelo2018conditional}, demonstrating that generalization over \sinet{} representations is at least equally as powerful as generalization over images.

%% file: sections/discussion.tex
The question of how to represent a signal is at the core of many problems across science and engineering. Implicit neural representations may provide a new tool for many of these by offering a number of potential benefits over conventional continuous and discrete representations. We demonstrate that periodic activation functions are ideally suited for representing complex natural signals and their derivatives using implicit neural representations. We also prototype several boundary value problems that our framework is capable of solving robustly. There are several exciting avenues for future work, including the exploration of other types of inverse problems and applications in areas beyond implicit neural representations, for example neural ODEs~\cite{chen2018neural}.

With this work, we make important contributions to the emerging field of implicit neural representation learning and its applications.	

%% file: sections/broader_impact.tex
The proposed \sinet{} representation enables accurate representations of natural signals, such as images, audio, and video in a deep learning framework. 
This may be an enabler for downstream tasks involving such signals, such as classification for images or speech-to-text systems for audio. 
Such applications may be leveraged for both positive and negative ends.
\sinet{} may in the future further enable novel approaches to the generation of such signals. 
This has potential for misuse in impersonating actors without their consent. For an in-depth discussion of such so-called DeepFakes, we refer the reader to a recent review article on neural rendering~\cite{tewari2020state}.

%% file: main.bbl
\begin{thebibliography}{53}
\providecommand{\natexlab}[1]{#1}
\providecommand{\url}[1]{\texttt{#1}}
\expandafter\ifx\csname urlstyle\endcsname\relax
  \providecommand{\doi}[1]{doi: #1}\else
  \providecommand{\doi}{doi: \begingroup \urlstyle{rm}\Url}\fi

\bibitem[Park et~al.(2019)Park, Florence, Straub, Newcombe, and
  Lovegrove]{park2019deepsdf}
Jeong~Joon Park, Peter Florence, Julian Straub, Richard Newcombe, and Steven
  Lovegrove.
\newblock Deepsdf: Learning continuous signed distance functions for shape
  representation.
\newblock \emph{Proc. CVPR}, 2019.

\bibitem[Mescheder et~al.(2019)Mescheder, Oechsle, Niemeyer, Nowozin, and
  Geiger]{mescheder2019occupancy}
Lars Mescheder, Michael Oechsle, Michael Niemeyer, Sebastian Nowozin, and
  Andreas Geiger.
\newblock Occupancy networks: Learning 3d reconstruction in function space.
\newblock In \emph{Proc. CVPR}, 2019.

\bibitem[Saito et~al.(2019)Saito, Huang, Natsume, Morishima, Kanazawa, and
  Li]{saito2019pifu}
Shunsuke Saito, Zeng Huang, Ryota Natsume, Shigeo Morishima, Angjoo Kanazawa,
  and Hao Li.
\newblock Pifu: Pixel-aligned implicit function for high-resolution clothed
  human digitization.
\newblock In \emph{Proc. ICCV}, pages 2304--2314, 2019.

\bibitem[Atzmon and Lipman(2020)]{atzmon2019sal}
Matan Atzmon and Yaron Lipman.
\newblock Sal: Sign agnostic learning of shapes from raw data.
\newblock In \emph{Proc. CVPR}, 2020.

\bibitem[Mildenhall et~al.(2020)Mildenhall, Srinivasan, Tancik, Barron,
  Ramamoorthi, and Ng]{mildenhall2020nerf}
Ben Mildenhall, Pratul~P Srinivasan, Matthew Tancik, Jonathan~T Barron, Ravi
  Ramamoorthi, and Ren Ng.
\newblock Nerf: Representing scenes as neural radiance fields for view
  synthesis.
\newblock \emph{arXiv preprint arXiv:2003.08934}, 2020.

\bibitem[Genova et~al.(2019{\natexlab{a}})Genova, Cole, Vlasic, Sarna, Freeman,
  and Funkhouser]{genova2019learning}
Kyle Genova, Forrester Cole, Daniel Vlasic, Aaron Sarna, William~T Freeman, and
  Thomas Funkhouser.
\newblock Learning shape templates with structured implicit functions.
\newblock In \emph{Proc. ICCV}, pages 7154--7164, 2019{\natexlab{a}}.

\bibitem[Genova et~al.(2019{\natexlab{b}})Genova, Cole, Sud, Sarna, and
  Funkhouser]{genova2019deep}
Kyle Genova, Forrester Cole, Avneesh Sud, Aaron Sarna, and Thomas Funkhouser.
\newblock Deep structured implicit functions.
\newblock \emph{arXiv preprint arXiv:1912.06126}, 2019{\natexlab{b}}.

\bibitem[Michalkiewicz et~al.(2019)Michalkiewicz, Pontes, Jack, Baktashmotlagh,
  and Eriksson]{michalkiewicz2019implicit}
Mateusz Michalkiewicz, Jhony~K Pontes, Dominic Jack, Mahsa Baktashmotlagh, and
  Anders Eriksson.
\newblock Implicit surface representations as layers in neural networks.
\newblock In \emph{Proc. ICCV}, pages 4743--4752, 2019.

\bibitem[Gropp et~al.(2020)Gropp, Yariv, Haim, Atzmon, and
  Lipman]{gropp2020implicit}
Amos Gropp, Lior Yariv, Niv Haim, Matan Atzmon, and Yaron Lipman.
\newblock Implicit geometric regularization for learning shapes.
\newblock \emph{arXiv preprint arXiv:2002.10099}, 2020.

\bibitem[Sitzmann et~al.(2019)Sitzmann, Zollh{\"o}fer, and
  Wetzstein]{sitzmann2019srns}
Vincent Sitzmann, Michael Zollh{\"o}fer, and Gordon Wetzstein.
\newblock Scene representation networks: Continuous 3d-structure-aware neural
  scene representations.
\newblock In \emph{Proc. NeurIPS}, 2019.

\bibitem[Jiang et~al.(2020)Jiang, Sud, Makadia, Huang, Nie{\ss}ner, and
  Funkhouser]{jiang2020local}
Chiyu Jiang, Avneesh Sud, Ameesh Makadia, Jingwei Huang, Matthias Nie{\ss}ner,
  and Thomas Funkhouser.
\newblock Local implicit grid representations for 3d scenes.
\newblock In \emph{Proc. CVPR}, pages 6001--6010, 2020.

\bibitem[Peng et~al.(2020)Peng, Niemeyer, Mescheder, Pollefeys, and
  Geiger]{peng2020convolutional}
Songyou Peng, Michael Niemeyer, Lars Mescheder, Marc Pollefeys, and Andreas
  Geiger.
\newblock Convolutional occupancy networks.
\newblock \emph{arXiv preprint arXiv:2003.04618}, 2020.

\bibitem[Chabra et~al.(2020)Chabra, Lenssen, Ilg, Schmidt, Straub, Lovegrove,
  and Newcombe]{chabra2020deep}
Rohan Chabra, Jan~Eric Lenssen, Eddy Ilg, Tanner Schmidt, Julian Straub, Steven
  Lovegrove, and Richard Newcombe.
\newblock Deep local shapes: Learning local sdf priors for detailed 3d
  reconstruction.
\newblock \emph{arXiv preprint arXiv:2003.10983}, 2020.

\bibitem[Chen and Zhang(2019)]{chen2019learning}
Zhiqin Chen and Hao Zhang.
\newblock Learning implicit fields for generative shape modeling.
\newblock In \emph{Proc. CVPR}, pages 5939--5948, 2019.

\bibitem[Oechsle et~al.(2019)Oechsle, Mescheder, Niemeyer, Strauss, and
  Geiger]{Oechsle2019ICCV}
Michael Oechsle, Lars Mescheder, Michael Niemeyer, Thilo Strauss, and Andreas
  Geiger.
\newblock Texture fields: Learning texture representations in function space.
\newblock In \emph{Proc. ICCV}, 2019.

\bibitem[Niemeyer et~al.(2020)Niemeyer, Mescheder, Oechsle, and
  Geiger]{Niemeyer2020CVPR}
Michael Niemeyer, Lars Mescheder, Michael Oechsle, and Andreas Geiger.
\newblock Differentiable volumetric rendering: Learning implicit 3d
  representations without 3d supervision.
\newblock In \emph{Proc. CVPR}, 2020.

\bibitem[Tewari et~al.(2020)Tewari, Fried, Thies, Sitzmann, Lombardi,
  Sunkavalli, Martin-Brualla, Simon, Saragih, Nie{\ss}ner,
  et~al.]{tewari2020state}
Ayush Tewari, Ohad Fried, Justus Thies, Vincent Sitzmann, Stephen Lombardi,
  Kalyan Sunkavalli, Ricardo Martin-Brualla, Tomas Simon, Jason Saragih,
  Matthias Nie{\ss}ner, et~al.
\newblock State of the art on neural rendering.
\newblock \emph{Proc. Eurographics}, 2020.

\bibitem[Niemeyer et~al.(2019)Niemeyer, Mescheder, Oechsle, and
  Geiger]{Niemeyer2019ICCV}
Michael Niemeyer, Lars Mescheder, Michael Oechsle, and Andreas Geiger.
\newblock Occupancy flow: 4d reconstruction by learning particle dynamics.
\newblock In \emph{Proc. ICCV}, 2019.

\bibitem[Kohli et~al.(2020)Kohli, Sitzmann, and Wetzstein]{kohli2020inferring}
Amit Kohli, Vincent Sitzmann, and Gordon Wetzstein.
\newblock Inferring semantic information with 3d neural scene representations.
\newblock \emph{arXiv preprint arXiv:2003.12673}, 2020.

\bibitem[Gallant and White(1988)]{Gallant:1988}
R.~Gallant and H.~White.
\newblock There exists a neural network that does not make avoidable mistakes.
\newblock In \emph{IEEE Int. Conference on Neural Networks}, pages 657--664,
  1988.

\bibitem[Zhumekenov et~al.(2019)Zhumekenov, Uteuliyeva, Kabdolov, Takhanov,
  Assylbekov, and Castro]{zhumekenov2019fourier}
Abylay Zhumekenov, Malika Uteuliyeva, Olzhas Kabdolov, Rustem Takhanov,
  Zhenisbek Assylbekov, and Alejandro~J Castro.
\newblock Fourier neural networks: A comparative study.
\newblock \emph{arXiv preprint arXiv:1902.03011}, 2019.

\bibitem[Sopena et~al.(1999)Sopena, Romero, and Alquezar]{sopena1999neural}
Josep~M Sopena, Enrique Romero, and Rene Alquezar.
\newblock Neural networks with periodic and monotonic activation functions: a
  comparative study in classification problems.
\newblock In \emph{Proc. ICANN}, 1999.

\bibitem[Wong et~al.(2002)Wong, Leung, and Chang]{wong2002handwritten}
Kwok-wo Wong, Chi-sing Leung, and Sheng-jiang Chang.
\newblock Handwritten digit recognition using multilayer feedforward neural
  networks with periodic and monotonic activation functions.
\newblock In \emph{Object recognition supported by user interaction for service
  robots}, volume~3, pages 106--109. IEEE, 2002.

\bibitem[Parascandolo et~al.(2016)Parascandolo, Huttunen, and
  Virtanen]{parascandolo2016taming}
Giambattista Parascandolo, Heikki Huttunen, and Tuomas Virtanen.
\newblock Taming the waves: sine as activation function in deep neural
  networks.
\newblock 2016.

\bibitem[Liu et~al.(2015{\natexlab{a}})Liu, Zeng, and
  Wang]{liu2015multistability}
Peng Liu, Zhigang Zeng, and Jun Wang.
\newblock Multistability of recurrent neural networks with nonmonotonic
  activation functions and mixed time delays.
\newblock \emph{IEEE Trans. on Systems, Man, and Cybernetics: Systems},
  46\penalty0 (4):\penalty0 512--523, 2015{\natexlab{a}}.

\bibitem[Koplon and Sontag(1997)]{koplon1997using}
Ren{\'e}e Koplon and Eduardo~D Sontag.
\newblock Using fourier-neural recurrent networks to fit sequential
  input/output data.
\newblock \emph{Neurocomputing}, 15\penalty0 (3-4):\penalty0 225--248, 1997.

\bibitem[Choueiki et~al.(1997)Choueiki, Mount-Campbell, and
  Ahalt]{choueiki1997implementing}
M~Hisham Choueiki, Clark~A Mount-Campbell, and Stanley~C Ahalt.
\newblock Implementing a weighted least squares procedure in training a neural
  network to solve the short-term load forecasting problem.
\newblock \emph{IEEE Trans. on Power systems}, 12\penalty0 (4):\penalty0
  1689--1694, 1997.

\bibitem[Alqu{\'e}zar~Mancho(1997)]{alquezar1997symbolic}
Ren{\'e} Alqu{\'e}zar~Mancho.
\newblock \emph{Symbolic and connectionist learning techniques for grammatical
  inference}.
\newblock Universitat Polit{\`e}cnica de Catalunya, 1997.

\bibitem[Sopena and Alquezar(1994)]{sopena1994improvement}
JM~Sopena and R~Alquezar.
\newblock Improvement of learning in recurrent networks by substituting the
  sigmoid activation function.
\newblock In \emph{Proc. ICANN}, pages 417--420. Springer, 1994.

\bibitem[Cand{\`e}s(1999)]{candes1999harmonic}
Emmanuel~J Cand{\`e}s.
\newblock Harmonic analysis of neural networks.
\newblock \emph{Applied and Computational Harmonic Analysis}, 6\penalty0
  (2):\penalty0 197--218, 1999.

\bibitem[Lin et~al.(2013)Lin, Guo, Cao, and Xu]{lin2013approximation}
Shaobo Lin, Xiaofei Guo, Feilong Cao, and Zongben Xu.
\newblock Approximation by neural networks with scattered data.
\newblock \emph{Applied Mathematics and Computation}, 224:\penalty0 29--35,
  2013.

\bibitem[Sonoda and Murata(2017)]{sonoda2017neural}
Sho Sonoda and Noboru Murata.
\newblock Neural network with unbounded activation functions is universal
  approximator.
\newblock \emph{Applied and Computational Harmonic Analysis}, 43\penalty0
  (2):\penalty0 233--268, 2017.

\bibitem[Stanley(2007)]{stanley2007compositional}
Kenneth~O Stanley.
\newblock Compositional pattern producing networks: A novel abstraction of
  development.
\newblock \emph{Genetic programming and evolvable machines}, 8\penalty0
  (2):\penalty0 131--162, 2007.

\bibitem[Mordvintsev et~al.(2018)Mordvintsev, Pezzotti, Schubert, and
  Olah]{mordvintsev2018differentiable}
Alexander Mordvintsev, Nicola Pezzotti, Ludwig Schubert, and Chris Olah.
\newblock Differentiable image parameterizations.
\newblock \emph{Distill}, 3\penalty0 (7):\penalty0 e12, 2018.

\bibitem[Klocek et~al.(2019)Klocek, Maziarka, Wo{\l}czyk, Tabor, Nowak, and
  {\'S}mieja]{klocek2019hypernetwork}
Sylwester Klocek, {\L}ukasz Maziarka, Maciej Wo{\l}czyk, Jacek Tabor, Jakub
  Nowak, and Marek {\'S}mieja.
\newblock Hypernetwork functional image representation.
\newblock In \emph{Proc. ICANN}, pages 496--510. Springer, 2019.

\bibitem[Lee and Kang(1990)]{lee1990neural}
Hyuk Lee and In~Seok Kang.
\newblock Neural algorithm for solving differential equations.
\newblock \emph{Journal of Computational Physics}, 91\penalty0 (1):\penalty0
  110--131, 1990.

\bibitem[Lagaris et~al.(1998)Lagaris, Likas, and
  Fotiadis]{lagaris1998artificial}
Isaac~E Lagaris, Aristidis Likas, and Dimitrios~I Fotiadis.
\newblock Artificial neural networks for solving ordinary and partial
  differential equations.
\newblock \emph{IEEE Trans. on neural networks}, 9\penalty0 (5):\penalty0
  987--1000, 1998.

\bibitem[He et~al.(2000)He, Reif, and Unbehauen]{he2000multilayer}
Shouling He, Konrad Reif, and Rolf Unbehauen.
\newblock Multilayer neural networks for solving a class of partial
  differential equations.
\newblock \emph{Neural networks}, 13\penalty0 (3):\penalty0 385--396, 2000.

\bibitem[Mai-Duy and Tran-Cong(2003)]{mai2003approximation}
Nam Mai-Duy and Thanh Tran-Cong.
\newblock Approximation of function and its derivatives using radial basis
  function networks.
\newblock \emph{Applied Mathematical Modelling}, 27\penalty0 (3):\penalty0
  197--220, 2003.

\bibitem[Sirignano and Spiliopoulos(2018)]{sirignano2018dgm}
Justin Sirignano and Konstantinos Spiliopoulos.
\newblock Dgm: A deep learning algorithm for solving partial differential
  equations.
\newblock \emph{Journal of Computational Physics}, 375:\penalty0 1339--1364,
  2018.

\bibitem[Raissi et~al.(2019)Raissi, Perdikaris, and
  Karniadakis]{raissi2019physics}
Maziar Raissi, Paris Perdikaris, and George~E Karniadakis.
\newblock Physics-informed neural networks: A deep learning framework for
  solving forward and inverse problems involving nonlinear partial differential
  equations.
\newblock \emph{Journal of Computational Physics}, 378:\penalty0 686--707,
  2019.

\bibitem[Berg and Nystr{\"o}m(2018)]{berg2018unified}
Jens Berg and Kaj Nystr{\"o}m.
\newblock A unified deep artificial neural network approach to partial
  differential equations in complex geometries.
\newblock \emph{Neurocomputing}, 317:\penalty0 28--41, 2018.

\bibitem[Chen et~al.(2018)Chen, Rubanova, Bettencourt, and
  Duvenaud]{chen2018neural}
Tian~Qi Chen, Yulia Rubanova, Jesse Bettencourt, and David~K Duvenaud.
\newblock Neural ordinary differential equations.
\newblock In \emph{Proc. NIPS}, pages 6571--6583, 2018.

\bibitem[P\'{e}rez et~al.(2003)P\'{e}rez, Gangnet, and Blake]{perez:2003}
Patrick P\'{e}rez, Michel Gangnet, and Andrew Blake.
\newblock Poisson image editing.
\newblock \emph{ACM Trans. on Graphics}, 22\penalty0 (3):\penalty0 313–318,
  2003.

\bibitem[Chen et~al.(2013)Chen, Cheng, Feng, and Wu]{chen2013optimal}
Zhongying Chen, Dongsheng Cheng, Wei Feng, and Tingting Wu.
\newblock An optimal 9-point finite difference scheme for the helmholtz
  equation with pml.
\newblock \emph{International Journal of Numerical Analysis \& Modeling},
  10\penalty0 (2), 2013.

\bibitem[Aghamiry et~al.(2019)Aghamiry, Gholami, and
  Operto]{aghamiry2019improving}
Hossein~S Aghamiry, Ali Gholami, and St{\'e}phane Operto.
\newblock Improving full-waveform inversion by wavefield reconstruction with
  the alternating direction method of multipliers.
\newblock \emph{Geophysics}, 84\penalty0 (1):\penalty0 R139--R162, 2019.

\bibitem[Van~Leeuwen and Herrmann(2013)]{van2013mitigating}
Tristan Van~Leeuwen and Felix~J Herrmann.
\newblock Mitigating local minima in full-waveform inversion by expanding the
  search space.
\newblock \emph{Geophysical Journal International}, 195\penalty0 (1):\penalty0
  661--667, 2013.

\bibitem[Boyd et~al.(2011)Boyd, Parikh, Chu, Peleato, Eckstein,
  et~al.]{boyd2011distributed}
Stephen Boyd, Neal Parikh, Eric Chu, Borja Peleato, Jonathan Eckstein, et~al.
\newblock Distributed optimization and statistical learning via the alternating
  direction method of multipliers.
\newblock \emph{Foundations and Trends{\textregistered} in Machine learning},
  3\penalty0 (1):\penalty0 1--122, 2011.

\bibitem[Liu et~al.(2015{\natexlab{b}})Liu, Luo, Wang, and
  Tang]{liu2015faceattributes}
Ziwei Liu, Ping Luo, Xiaogang Wang, and Xiaoou Tang.
\newblock Deep learning face attributes in the wild.
\newblock In \emph{Proc. ICCV}, December 2015{\natexlab{b}}.

\bibitem[Garnelo et~al.(2018)Garnelo, Rosenbaum, Maddison, Ramalho, Saxton,
  Shanahan, Teh, Rezende, and Eslami]{garnelo2018conditional}
Marta Garnelo, Dan Rosenbaum, Chris~J Maddison, Tiago Ramalho, David Saxton,
  Murray Shanahan, Yee~Whye Teh, Danilo~J Rezende, and SM~Eslami.
\newblock Conditional neural processes.
\newblock \emph{arXiv preprint arXiv:1807.01613}, 2018.

\bibitem[Eslami et~al.(2018)Eslami, Rezende, Besse, Viola, Morcos, Garnelo,
  Ruderman, Rusu, Danihelka, Gregor, et~al.]{eslami2018neural}
SM~Ali Eslami, Danilo~Jimenez Rezende, Frederic Besse, Fabio Viola, Ari~S
  Morcos, Marta Garnelo, Avraham Ruderman, Andrei~A Rusu, Ivo Danihelka, Karol
  Gregor, et~al.
\newblock Neural scene representation and rendering.
\newblock \emph{Science}, 360\penalty0 (6394):\penalty0 1204--1210, 2018.

\bibitem[Kim et~al.(2019)Kim, Mnih, Schwarz, Garnelo, Eslami, Rosenbaum,
  Vinyals, and Teh]{kim2019attentive}
Hyunjik Kim, Andriy Mnih, Jonathan Schwarz, Marta Garnelo, Ali Eslami, Dan
  Rosenbaum, Oriol Vinyals, and Yee~Whye Teh.
\newblock Attentive neural processes.
\newblock \emph{Proc. ICLR}, 2019.

\bibitem[Ha et~al.(2017)Ha, Dai, and Le]{ha2016hypernetworks}
David Ha, Andrew Dai, and Quoc~V Le.
\newblock Hypernetworks.
\newblock In \emph{Proc. ICLR}, 2017.

\end{thebibliography}
